\documentclass[conference]{IEEEtran}
\IEEEoverridecommandlockouts

\usepackage{cite}
\usepackage{amsmath,amssymb,amsfonts}
\usepackage{algorithmic}
\usepackage{graphicx}
\usepackage{textcomp}
\usepackage{xcolor}
\usepackage{booktabs}
\usepackage{multirow}
\usepackage{subfig}
\usepackage{soul}

\usepackage[colorinlistoftodos,prependcaption,textsize=tiny]{todonotes}
\newcommand{\comment}[1]{}
\usepackage{comment}
\usepackage[table]{xcolor} 
\usepackage{subcaption}
\usepackage{array}          
\usepackage{tabularx}      

\newcommand\copyrighttext{%
  \footnotesize \textcopyright 2025 IEEE. Personal use of this material is permitted.  Permission from IEEE must be obtained for all other uses, in any current or future media, including reprinting/republishing this material for advertising or promotional purposes, creating new collective works, for resale or redistribution to servers or lists, or reuse of any copyrighted component of this work in other works.
  }
\newcommand\copyrightnotice{%
\begin{tikzpicture}[remember picture,overlay]
\node[anchor=south,yshift=10pt] at (current page.south) {\fbox{\parbox{\dimexpr\textwidth-\fboxsep-\fboxrule\relax}{\copyrighttext}}};
\end{tikzpicture}%
}

\newcolumntype{P}[1]{>{\centering\arraybackslash}p{#1}}

\def\BibTeX{{\rm B\kern-.05em{\sc i\kern-.025em b}\kern-.08em
    T\kern-.1667em\lower.7ex\hbox{E}\kern-.125emX}}

\begin{document}

\title{Physics-guided Neural Network-based Shaft Power Prediction for Vessels\\
}

\author{\IEEEauthorblockN{Dogan Altan, Hamza Haruna Mohammed, Glenn Terje Lines, Dusica Marijan}
\IEEEauthorblockA{\textit{Simula Research Laboratory} \\
Oslo, Norway\\
\{dogan, hamzahm, glennli, dusica\}@simula.no}
\and
\IEEEauthorblockN{Arnbjørn Maressa}
\IEEEauthorblockA{\textit{Navtor AS} \\
Egersund, Norway\\
}
}
\maketitle
\copyrightnotice

\begin{abstract}
Optimizing maritime operations, particularly fuel consumption for vessels, is crucial, considering its significant share in global trade. As fuel consumption is closely related to the shaft power of a vessel, predicting shaft power accurately is a crucial problem that requires careful consideration to minimize costs and emissions. Traditional approaches, which incorporate empirical formulas, often struggle to model dynamic conditions, such as sea conditions or fouling on vessels. In this paper, we present a hybrid, physics-guided neural network-based approach that utilizes empirical formulas within the network to combine the advantages of both neural networks and traditional techniques. We evaluate the presented method using data obtained from four similar-sized cargo vessels and compare the results with those of a baseline neural network and a traditional approach that employs empirical formulas. The experimental results demonstrate that the physics-guided neural network approach achieves lower mean absolute error, root mean square error, and mean absolute percentage error for all tested vessels compared to both the empirical formula-based method and the base neural network.
\end{abstract}


\section{Introduction}

Maritime transport is central to global trade, moving more than 80\% of all goods worldwide by volume, and thus playing a vital role in the global economy. Given its immense scale, optimization in maritime operations has significant potential to yield substantial economic and environmental benefits. For instance, reducing fuel consumption through accurate shaft power prediction directly translates to lower operational costs for shipping companies and a significant decrease in greenhouse gas emissions, aligning with global efforts to combat climate change \cite{adland2018energy}.

Reducing emissions from maritime transport is crucial not only for mitigating climate change but also for ensuring the long-term economic and operational sustainability of the global shipping industry. The International Maritime Organization’s 2023 greenhouse gas (GHG) reduction strategy emphasizes the need to reduce emissions from ships in line with international climate goals \cite{mepc20232023}. A significant portion of these emissions stems from fuel consumption, which is heavily influenced by factors such as calm water resistance, wind, and wave effects \cite{coraddu2019data,lang2021practical}. Fouling on vessels, for instance, increases hydrodynamic resistance and can degrade energy efficiency by over 10\% if not correctly managed \cite{coraddu2019data}. Meanwhile, advancements in machine learning and hybrid modeling techniques enable more accurate predictions of ship fuel consumption under diverse environmental conditions, allowing for optimized voyage planning and reduced emissions \cite{fan2022joint}. As trade volumes grow, failure to address these inefficiencies could escalate both fuel costs and environmental impacts, thus reinforcing the need for integrated, data-driven strategies to cut emissions while preserving maritime competitiveness \cite{ferrari2023impact, yan2024improving}.

Several studies have explored various data-driven models to predict shaft power and fuel consumption \cite{kim2024interpretable}, considering factors such as hull fouling and environmental conditions (i.e., weather conditions \cite{parkes2019efficient}), highlighting the complexity and importance of accurate predictions for operational efficiency \cite{bakka2023estimating} \cite{karagiannidis2021data}. However, such models typically rely on empirical regressions or statistical methods that cannot generalize well across varying ship types, sea states, and data quality scenarios \cite{gkerekos2019machine}, \cite{fan2022review}. A fundamental issue is that these approaches often produce predictions that violate basic marine engineering principles \cite{wang2023innovative}, particularly when faced with dynamic operational conditions or data sparsity \cite{mylonopoulos2023comprehensive}, \cite{yan2021data}.

These shortcomings become particularly apparent when models are applied to real-world operational scenarios. For instance, statistical approaches often struggle with the inherent noise and variability in noon report data \cite{bayraktar2024marine}. At the same time, pure machine learning models may generate physically impossible fuel consumption values under certain conditions \cite{koh2018buoyancy}. The limitations are further compounded when attempting to account for dynamic factors, such as sudden weather changes or progressive hull fouling \cite{coraddu2019data}. Therefore, hybrid approaches, such as Physics-Guided Neural Networks (PGNNs) \cite{daw2017physics} and Physics-Informed Neural Networks (PINNs) \cite{raissi2019physics}, emerge as a way to overcome the limitations of these individual approaches. Such hybrid approaches can incorporate domain-specific physical constraints, such as hydrodynamic resistance, propulsion equations, and energy conservation, directly into the learning process. By aligning predictions with known physical behaviors and dynamics, PGNNs and PINNs offer a more accurate and generalizable approach to fuel consumption or shaft power prediction, making them ideal for supporting decarbonization efforts in line with the IMO’s GHG reduction strategy \cite{mepc20232023}.

In this paper, we present a physics-guided neural network-based method for tackling the shaft power prediction problem in maritime applications. The method incorporates several resistance factors, including calm water, wind, and waves, into the model to guide the training process. Our proposed method also employs a polynomial model to predict shaft rotational speed (in revolutions per minute (RPM)) and uses these predictions as a feature to predict shaft power. The contributions of this work are summarized as follows:

\begin{itemize}
    \item We present a physics-guided neural network model to predict vessel shaft power, where physical formulas are incorporated into the model's loss function to include domain-specific insights directly into the learning process.   

    \item Beyond calm water resistance, the network's training process is informed by the inclusion of wave and wind-related resistance components, ensuring a more complete representation of environmental factors.

    \item We evaluate the presented method using real-world vessel data obtained from similar-sized cargo vessels and compare it with baselines. We further analyze the performance of the proposed method under severe sea conditions (i.e., high waves).
\end{itemize}

This paper is organized as follows. First, we review the literature on the shaft power prediction problem. We then present our shaft power prediction method, based on physics-guided neural networks, followed by the evaluation of the presented method using data obtained from vessels. Finally, we conclude the paper with concluding remarks and potential directions. 

\section{Related Work}

The problem of shaft power prediction has been investigated in the literature by various researchers from different perspectives. This section summarizes the related work on shaft power prediction.

\begin{figure*}[!htbp]
  \centering
  \includegraphics[width=0.8\textwidth]{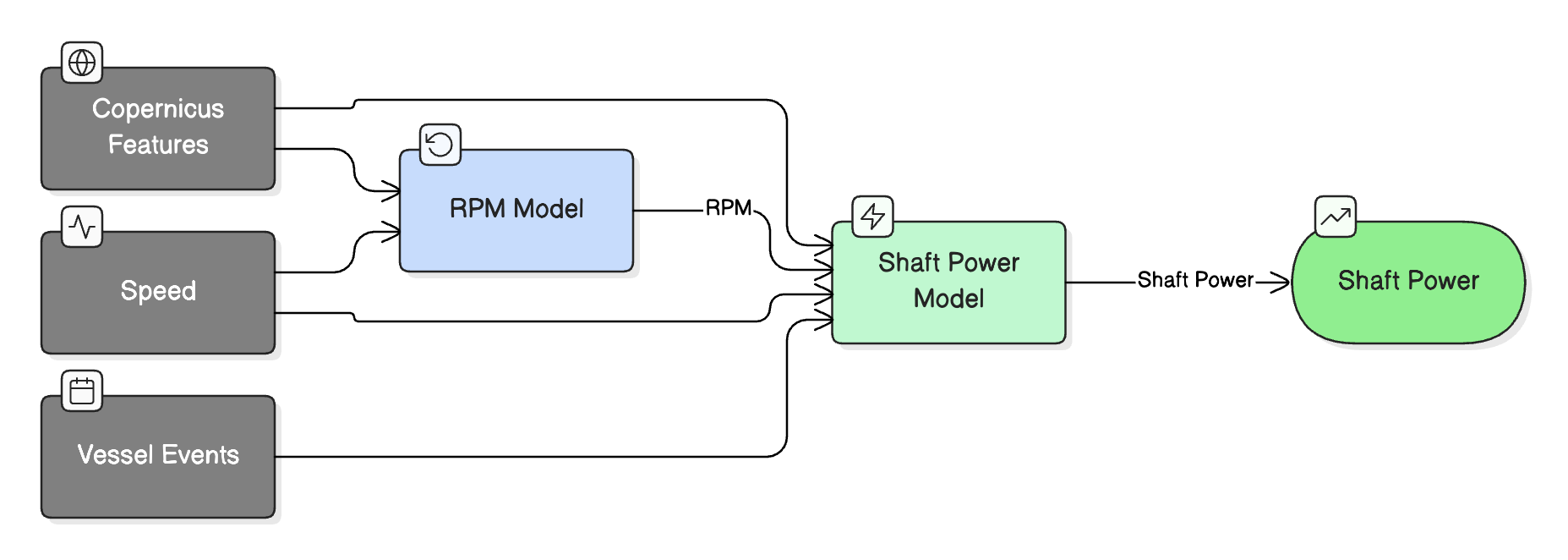} 
  \caption{The presented two-stage shaft power prediction model.}
  \label{fig:model}
\end{figure*}

Maintenance events, such as hull cleaning and dry-docking, have been shown to significantly impact a vessel's fuel consumption \cite{adland2018energy}, due to marine fouling on the vessel's hull and propeller. One study investigates and combines genetic algorithms with long short-term memory (LSTM) models to assess vessel speed, considering the degradation of hull and propeller performance \cite{huang2024assessment}. The presented two-stage optimization algorithm selects relevant features and parameters with a genetic algorithm, followed by an LSTM-based learning approach. Another work investigates an Unsteady Reynolds Averaged Navier–Stokes (URANS)-based method to measure the effect of biofouling on ship self-propulsion performance \cite{song2020penalty}. Another study \cite{uzun2019time} investigates biofouling as a time-dependent growth model for predicting fouling impact on a ship's resistance.

Various methods have also been studied to predict a vessel's shaft power while accounting for the effects of fouling. In one study \cite{bakka2023estimating}, a Bayesian generalized additive model is presented, where fouling is modeled as an additive factor for shaft power prediction, alongside other additive factors such as speed and wave conditions. Yet another work models the shaft power prediction problem as a linear regression problem \cite{kim2024interpretable}. The model comprises three primary factors for predicting shaft power: RPM, fouling effects, and environmental conditions, excluding speed.

Another work investigates four different machine learning approaches to predict shaft power, considering fouling \cite{laurie2021machine}. K-nearest neighbours (KNN), decision trees, random forests, and artificial neural networks are implemented to process speed through water, relative wind speed, sea temperature, trim, and draft features. The number of days that have elapsed after the cleaning event is considered in the model to capture fouling effects. In another work \cite{karagiannidis2021data}, feed-forward neural networks (FNNs) are investigated for predicting a vessel's shaft power and fuel consumption after applying preprocessing methods, including an imputation algorithm. Feature generation has also been employed, including the introduction of a variable for fouling and accounting for the time elapsed from the first timestamp as a logarithmic function. Unlike these studies, we individually incorporate vessel maintenance events into the model and analyze them accordingly, rather than aggregating them into a single variable. In addition, another study investigates the impact of different neural network layer configurations on the performance of shaft power prediction \cite{parkes2018physics}.

A study presents a neural network-based method, grounded in physics, to predict the main power of the vessel engine \cite{bourchas2025physics}. The propeller law is incorporated into the network's loss function as a derivative term, in addition to the mean squared error loss for prediction. Unlike this work, we also include the resistance caused by wind and waves in the physical loss component. In another work \cite{zhou2024novel}, the fuel consumption of a vessel is predicted using a method that combines Gaussian processes with quantile regression theory. Ship resistance caused by the wind and waves is also taken into account to predict the shaft power of a vessel. In another study \cite{kim2023modelling}, the resistance of a ship, including calm water, wave, and wind resistances, is calculated with data pre-processing, and the shaft power is estimated using the resistance formulas. A transfer learning-based approach is also proposed to predict vessel shaft power, and a base model is trained with synthetic data obtained from physics-based resistance simulations and then fine-tuned with real-world data \cite{mavroudis2025application}. Unlike this work, our approach does not require the use of synthetic data, and we directly incorporate vessel resistance into the loss function of the model.

In contrast to earlier work, we incorporate all three resistance components (calm water, waves, and winds) directly into our neural network architecture through physical loss constraints, creating a unified framework that combines the interpretability of empirical formulas with the flexibility of deep learning. While previous studies either focused on isolated resistance types \cite{bourchas2025physics} or used separate preprocessing stages \cite{kim2023modelling}, our Physics-Guided Neural Network (PGNN)-based approach jointly optimizes physical consistency and data-driven patterns in an end-to-end manner, achieving superior generalization across operational conditions. This hybrid approach advances beyond propeller-only physics and decoupled resistance modeling by embedding the complete hydrodynamic system directly into the learning objective.
\vspace{-0.09cm}
\section{Physics-guided Neural Network-based Shaft Power Prediction}

In this section, we introduce the data and features used in this study. We then present the empirical formulas related to the resistance of a vessel. Later, we present the shaft power prediction method, incorporating its physics aspect. 
\vspace{-0.09cm}
\subsection{Data}

We use a dataset that includes data from four vessels (Vessel A-D), all of which are cargo vessels with similar lengths, ranging from 204 meters (Vessel A) to 208 meters (Vessel B-D). The sensor data is sampled every 15 minutes. As the dataset is confidential, we anonymize the vessel names and present them with the corresponding labels. Table \ref{tab:dataset} presents the intervals and the number of data instances in the datasets for each vessel. We also provide the dry docking event dates for each vessel. In addition to the dry docking event, the vessels undergo a propeller polishing event periodically, approximately twice a year.

\begin{table*}[h]
    \centering
    \caption{Number of instances and data periods for each vessel in training and test sets, including dry docking dates.}
    \renewcommand{\arraystretch}{1.5}
    \begin{tabular}{|P{1.5cm}|P{2.4cm}|P{2.2cm}|P{3.4cm}|P{3.4cm}|P{2.5cm}|}
        \rowcolor{gray!20}
        \hline
        \textbf{Vessel} & \textbf{Train Set Instances} & \textbf{Test Set Instances} & \textbf{Train Dates (Start/End)} & \textbf{Test Dates (Start/End)} & \textbf{Dry Docking Date} \\
        \hline
        Vessel A & 12787 & 17827 & 01-01-2022 / 31-12-2022 & 01-01-2023 / 31-12-2023 & 09-11-2022 \\
        \hline
        Vessel B & 27213 & 10757 & 01-01-2023 / 31-05-2024 & 01-06-2024 / 08-08-2024 & 23-05-2023 \\
        \hline
        Vessel C & 28013 & 11218 & 01-01-2023 / 31-05-2024 & 01-06-2024 / 08-08-2024 & 09-11-2023 \\
        \hline
        Vessel D & 18887 & 9945 & 01-01-2023 / 31-12-2023 & 01-01-2024 / 08-08-2024 & 14-08-2023 \\
        \hline
    \end{tabular}
    \label{tab:dataset}
\end{table*}

We use the following features to predict vessel shaft power: speed through water ($V$), draught ($T$), sea depth ($depth_\text{sea}$), sea temperature ($t_\text{sea}$), wave height ($h_\text{wave}$), swell height ($h_\text{swell}$), wave direction ($d_\text{wave}$), swell direction ($d_\text{swell}$), wind direction ($d_\text{wind}$) and wind speed ($v_\text{wind}$). Inspired by a previous study \cite{laurie2021machine}, we include vessel maintenance event-related features in the network, but as a separate feature for each event. We incorporate the number of days elapsed after the propeller polish ($days_{p}$) and dry docking ($days_{d}$) events. Note that all direction-based features are relative to the vessel. In addition to these features, we also employ a separate model to predict shaft rotational speed (RPM), which we use as an input to the neural network. Wave and wind-related features are obtained from Copernicus \cite{CopernicusEU}. Figure \ref{fig:model} depicts the general overview of the two-stage shaft power prediction method.

\subsection{Empirical Formulas (EF) for Resistance}
\label{trad-model-section}
The empirical formulas that we use take into account three types of resistance caused by different sources for the ship: calm water, wind and waves. These formulas are inspired by and built upon the recommended procedures and guidelines published by the International Towing Tank Conference for preparation, conduct and analysis of speed/power trials \cite{ITTCSpeedPower}, and are significantly adapted for our work.

\subsubsection{Calm water resistance} Calm water resistance is the force that is induced by still water due to the interaction between a ship's hull and the water. The propulsion system of the ship primarily generates power to counteract this force, allowing it to sail at the desired speed. It is formulated as follows:

\begin{equation}
    R_{\text{Calm}} = \frac{c\cdot(a + T)\cdot(b + V)^3}{V}
    \label{calm_equa}
\end{equation}

In this formula, $R_\text{Calm}$ is the calm resistance, $a$, $b$ and $c$ are coefficients to be learned, \textit{T} is the draught of the ship, and \textit{V} is speed through water ($V > 0$).

\subsubsection{Wind resistance}  

Wind resistance is the force the air applies to the vessel (i.e., the vessel portion above the waterline), acting in opposition to the vessel’s movement, and is calculated with the following formulas.

First, the air density ($\rho$) is calculated from the air temperature ($t_{\text{air}}$). 

\begin{equation}
    \rho = (1 + t_{\text{air}}/273.15)^{-1}  
\end{equation}

Then, the following is calculated as a dependent variable to account for the drag force by wind:

\begin{equation}
\begin{split}
C_x = \rho \cdot \Big( &f_\text{c} \cdot f_\text{h} 
+ (1 - f_\text{c}) \\
&\cdot (f_\text{h} \cdot |\cos(d_\text{wind})| + f_\text{s} \cdot |\sin(d_\text{wind})|) \Big)
\end{split}
\end{equation}

In this formula, $f_c$, $f_h$ and $f_s$ are cylinder, headwind and sidewind factors, respectively, and learned from the training data. Then, the wind resistance ($R_{wind}$) is calculated as follows:  

\begin{equation}
    R_{Wind} = C_x \cdot v_\text{wind}^2 \cdot \cos(d_\text{wind})
\end{equation}

 \begin{figure*}[!htbp]
  \centering
  \includegraphics[width=1.05\textwidth]{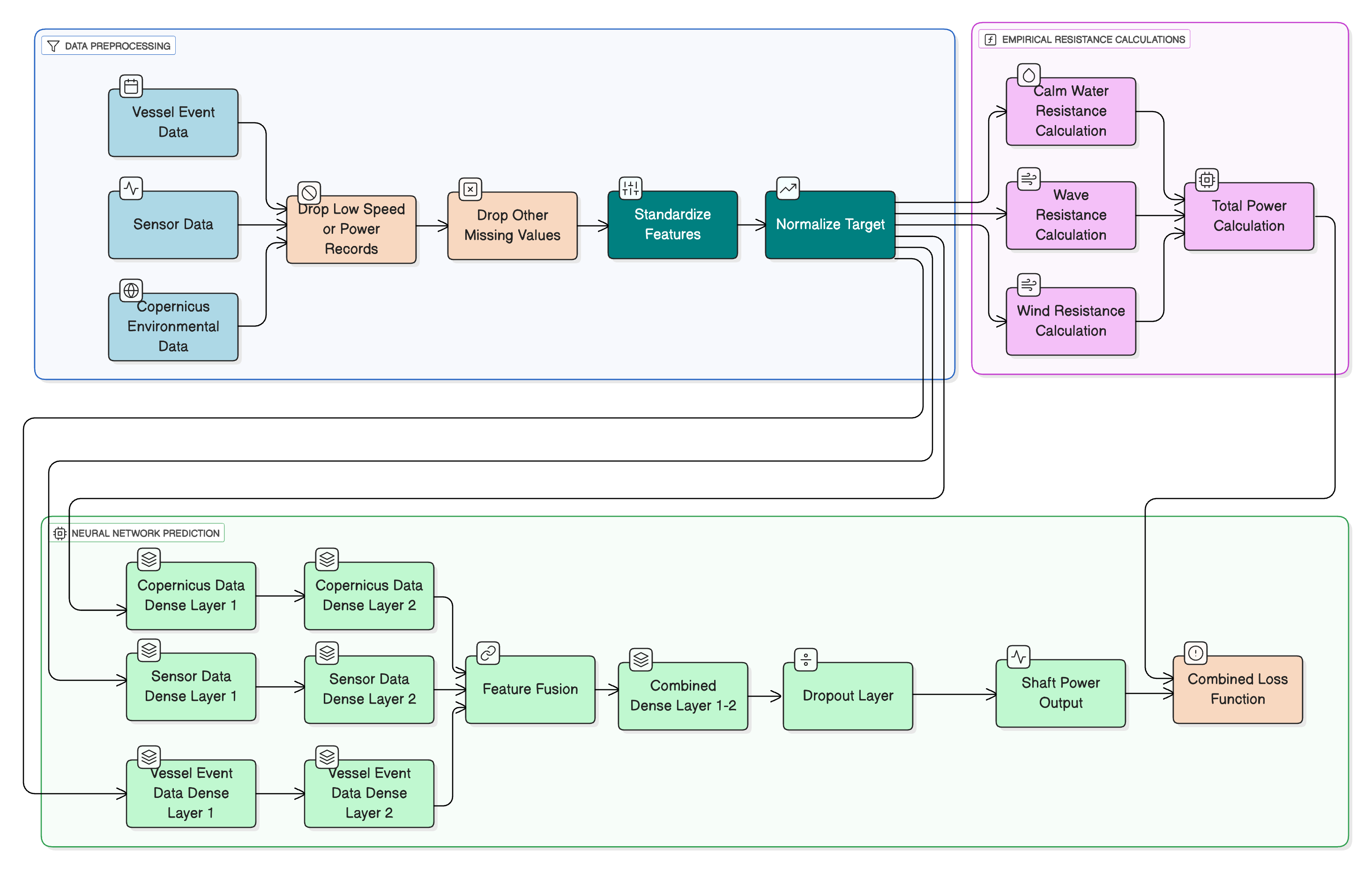} 
  \caption{Physics-guided Neural Network Pipeline for Shaft Power Prediction.}
  \label{fig:model-detailed}
\end{figure*}

\subsubsection{Wave resistance}
Wave resistance refers to the type of drag encountered by a vessel caused by the wind waves or swell. It is linked to the total resistance of the ship, thereby making it essential for fuel consumption and shaft power. Wave resistance for a ship is calculated as follows. First, the head wave resistance is calculated.

\begin{equation}
    R_\text{wave}^{\text{head}} = h_\text{wave}^\gamma \cdot f_g \cdot d_{\text{water}}
    \label{wave_resistance_head_equation}
\end{equation}

In this formula, $\gamma$ is a constant and is set empirically to 2.5 in our work. $f_g$ is the geometric factor that is learned from the data, and $d_\text{water}$ is the water density. Finally, the wave resistance is calculated. 

\begin{equation}
    R_{\text{Wave}} = R_\text{wave}^{\text{head}} \cdot (0.667 + 0.333 \cdot \cos(d_\text{wave}))
    \label{wave_resitance_equation}
\end{equation}

Considering all resistance components that are presented, the required power of a vessel can be calculated as follows:

\begin{equation}
    P_{\text{Physical}} = (R_{\text{Calm}} + R_{\text{Wind}} + R_{\text{Wave}}) \cdot V 
    \label{trad-model-power-equation}
\end{equation}

To learn the parameters in the presented formulas, we fit the measured shaft power values in the training set to Formula \ref{trad-model-power-equation} by minimizing the mean squared error, thereby obtaining the coefficients in the resistance formulas. Then, we use the learned parameters to calculate predicted shaft power for the empirical formula-based method for the test set. Note that the units for the variables in the formulas are not specified here, as they are arbitrary and canceled out in the calculations.

\subsection{Polynomial Models for RPM Prediction}

The RPM is modeled using a multiplicative function of univariate polynomials:
\[ \mbox{RPM} (u) =  \prod_{i=1}^n p_i(u_i) \]
where $n$ is the number of used features, $u$ is a vector of all features ordered such that the used ones are listed first,
and is $p_i$ the polynomial used for feature number $i$.
Features are chosen manually by first picking $u_1$ = $V$ (speed through water), 
then iteratively picking additional features based on what correlated the most with the residual error, 
until no significant improvement is observed. This gives a total of four ($n=4$) features: $V$, $T$ (draught),
$v_\text{wind}$ and $h_\text{swell}$. The order is set to 3 for all polynomials.

\subsection{Neural Networks for Shaft Power Prediction}

We utilize neural networks, which employ physical formulas presented in Section \ref{trad-model-section}, to provide insights into the model and predict vessel shaft power. Our neural network design consists of dense layers, and each feature type (features obtained from Copernicus, sensors, and external sources (i.e., vessel event features)) is first processed with two dense layers. The Copernicus features are processed within dense layers with 128 and 64 neurons, and the rest of the features are processed within 64 and 32 neurons. Then, the outputs are concatenated with dense layers of sizes 128 and 64. Later, a dropout layer is applied with a probability of 0.2, and shaft power is predicted using a final dense layer. Figure \ref{fig:model-detailed} depicts the detailed overview of the presented shaft power prediction method.

To incorporate physical constraints into the model, we modify the loss function used during training. In addition to the regular calculated loss (the difference between the model predictions and the ground truth values), the physical loss is also considered.
Specifically, we compute the difference between the model predictions and the shaft power values estimated using Formula \ref{trad-model-power-equation}. This physical loss is then scaled by a weighting factor ($\lambda$). The total loss function is defined as the sum of the standard prediction loss and the weighted physical loss (multiplied by  $\lambda$).


\section{Experiments}
In this section, we first present the experimental setup for the conducted experiments. We then present the experimental evaluation of the physics-guided shaft power prediction method.
\label{experiment_section}

\subsection{Experimental Setup}

\subsubsection{Training}

We preprocess the data before training by removing instances with speeds less than 5 knots and shaft powers less than 500 kilowatts (kW). We also drop data instances with missing values from the dataset.

We use the Adam optimizer in training and calculate the loss using the mean absolute error. Before training, we standardize the input features and normalize the target feature, which is shaft power. We set the batch size to 16, and use 20\% of the training set for validation. We employ an early stopping scheme and set the patience to 10.

To determine the $\lambda$ coefficients for the experiments, we train the models using $\lambda$ values ranging from 0.05 to 1 and select the configuration that achieves the best performance on the test set. As a result, the $\lambda$ values used for physical loss calculation were set to 0.1, 0.1, 0.05, and 0.4 for Vessels A, B, C, and D, respectively.

\subsubsection{Research Questions}

We investigate the following research questions (RQ) in the experiments to evaluate the proposed shaft power prediction algorithm.

\begin{itemize}
    \item \textbf{RQ1:} How does the presented physics-guided method compare with the baselines to predict the power of the vessel shaft?

    \item \textbf{RQ2:} How do the predictions of the physics-guided method compare to the ground truth over time for each vessel?

    \item \textbf{RQ3:} How does the proposed method perform under challenging sea conditions compared to the baselines?
    
\end{itemize}

\begin{table*}[ht]
    \centering
    \caption{Performance analysis (mean ± std) of different methods across vessels. \textbf{EF:} Empirical Formulas, \textbf{NN:} Neural Network, \textbf{PGNN:} Physics-guided Neural Network.}
    \renewcommand{\arraystretch}{2.0} 
    \rowcolors{2}{gray!10}{white}
    \begin{tabularx}{\textwidth}{ p{2.0cm} p{1.3cm} >{\centering\arraybackslash}X >{\centering\arraybackslash}X >{\centering\arraybackslash}X >{\centering\arraybackslash}X }
        \toprule
        \rowcolor{gray!30}
        \textbf{Vessel} & \textbf{Method} & \textbf{MAE} & \textbf{RMSE} & \textbf{R\textsuperscript{2}} & \textbf{MAPE (\%)} \\
        \midrule
        
        \multirow{3}{*}{A}
                           & \textbf{EF}    & 1450.10 ± 17.39 & 1685.93 ± 21.51 & -0.483 ± 0.03 & 26.28 ± 0.00 \\
        \textbf{Vessel A}  & \textbf{NN}    & 837.14 ± 134.06 & 1064.66 ± 167.84 & 0.395 ± 0.19 & 15.80 ± 2.44 \\
                           & \textbf{PGNN}  & 761.12 ± 109.69 & 972.52 ± 130.46 & 0.498 ± 0.12 & 14.50 ± 2.33 \\
        
        \midrule
        
        \multirow{3}{*}{B}
                          & \textbf{EF}    & 772.09 ± 6.44 & 862.32 ± 6.20 & -2.307 ± 0.08 & 29.03 ± 0.00 \\
        \textbf{Vessel B} & \textbf{NN}    & 300.62 ± 26.31 & 447.04 ± 32.97 & 0.105 ± 0.12 & 11.80 ± 0.98 \\ 
                          & \textbf{PGNN}  & 268.54 ± 13.33 & 428.54 ± 29.70 & 0.182 ± 0.07 & 10.90 ± 1.25 \\
        
        \midrule
        
        \multirow{3}{*}{C}
                          & \textbf{EF}    & 1036.35 ± 3.03 & 1101.58 ± 3.42 & 0.011 ± 0.00 & 36.30 ± 0.00 \\
        \textbf{Vessel C} & \textbf{NN}    & 628.37 ± 148.33 & 763.78 ± 151.44 & 0.505 ± 0.20 & 22.40 ± 5.39 \\ 
                          & \textbf{PGNN}  & 582.18 ± 105.49 & 708.21 ± 108.25 & 0.581 ± 0.12 & 20.90 ± 4.44 \\
        
        \midrule
        
        \multirow{3}{*}{D}
                          & \textbf{EF}    & 799.83 ± 7.63 & 1239.55 ± 12.42 & -0.195 ± 0.04 & 23.25 ± 0.00 \\
        \textbf{Vessel D} & \textbf{NN}    & 819.80 ± 113.03 & 951.49 ± 108.24 & 0.287 ± 0.16 & 17.00 ± 2.00 \\
                          & \textbf{PGNN}  & 775.68 ± 125.31 & 906.81 ± 117.63 & 0.349 ± 0.17 & 16.20 ± 2.27 \\ 
        
        \bottomrule
    \end{tabularx}
    \label{tab:perf-comp}
\end{table*}

\subsubsection{Evaluation Metrics}
We use the following performance metrics to evaluate the proposed shaft power prediction method. 

\begin{itemize}
    \item Mean absolute error (MAE): MAE calculates the average absolute difference of a regression model between its predictions and actual (ground truth) values. MAE is calculated with the following formula. 

\begin{equation*}
\text{MAE} = \frac{1}{n} \sum_{i=1}^{n} |y_i - \hat{y}_i|
\end{equation*}

In the formula, $\hat{y}_i$ is the predicted value by the model, $y_i$ is the ground truth, and $n$ is the number of data instances (i.e., predictions).

    \item Root mean square error (RMSE): Unlike MAE, the difference between the predicted and actual values is squared to assign relatively high weights to significant errors, and then the summation of all errors is square-rooted. The following is used to calculate RMSE. 

\begin{equation*}
\text{RMSE} = \sqrt{\frac{1}{n} \sum_{i=1}^{n} (y_i - \hat{y}_i)^2}
\end{equation*}

    \item Mean absolute percentage error (MAPE): MAPE stands for the mean of the percentage error of the predictions output by a model, and can be calculated as follows. 
\begin{equation*}
\text{MAPE} = \frac{1}{n} \sum_{i=1}^{n} \left| \frac{y_i - \hat{y}_i}{y_i} \right| \times 100\%
\end{equation*}
    \item Coefficient of determination (R$^2$): R$^2$, also named as coefficient of determination, measures how well the model's predictions approximate the actual data.
\begin{equation*}
R^2 = 1 - \frac{\sum_{i=1}^{n} (y_i - \hat{y}_i)^2}{\sum_{i=1}^{n} (y_i - \bar{y})^2}
\end{equation*}
In the formula, $\bar{y}$ denotes the mean of the actual values.
\end{itemize}

\subsection{Experimental Evaluation}

\subsubsection{RQ1 Comparison with Baselines} In this experiment, we evaluate the proposed method's performance and compare it with baselines. We compare our proposed method with two baselines. The first baseline calculates the vessel shaft power with the empirical formulas presented in Section \ref{trad-model-section}. The other baseline is the presented shaft power prediction method, which does not utilize physical formulas in the loss function and employs a mean absolute error loss as the difference between the predictions and the ground truth.

Table \ref{tab:perf-comp} presents the performance analysis of the proposed method and a comparison of the proposed method with the baselines. The first column in the table corresponds to the vessel name, and the next column corresponds to the method. For brevity purposes, we use abbreviations for the methods: EF denotes the method that uses empirical formulas to calculate shaft power, NN denotes the neural network that does not consider empirical formulas as physical constraints in the loss function, and PGNN denotes our proposed method. The remaining columns correspond to the performance metrics. As we repeat each experiment 10 times, we report the mean and standard deviation scores for each metric. 

As shown in the table, PGNN outperforms EF across all metrics while reducing errors (MAE, RMSE, and MAPE). Compared to NN, PGNN provides lower error scores and a higher $R^2$ score for Vessel A, B, C and D, as well. Upon further investigation of the scores for Vessel D, it is observed that for MAE, EF yields a better score than NN, while NN achieves better scores than EF for the other metrics. It is also worth noting that EF provides either close to zero (Vessel C) or negative (Vessels A, B, and D) $R^2$ scores to predict shaft power, indicating poor performance in fitting the data. Note that our method does not use ground truth RPM to provide a fair comparative analysis between the methods, as EF do not employ RPM. Therefore, our proposed method first predicts RPM using a polynomial model and then uses the predicted RPM values as a feature. The average MAPE for the predicted RPM values with the polynomial model is 3.95\%. One should also note that excluding the predicted RPM as a feature in the presented method yields an approximate decrease of 1.6\% in the MAPE.

\begin{figure*}[!t]
\centering
\subfloat[Vessel A]{\includegraphics[width=0.50\textwidth]{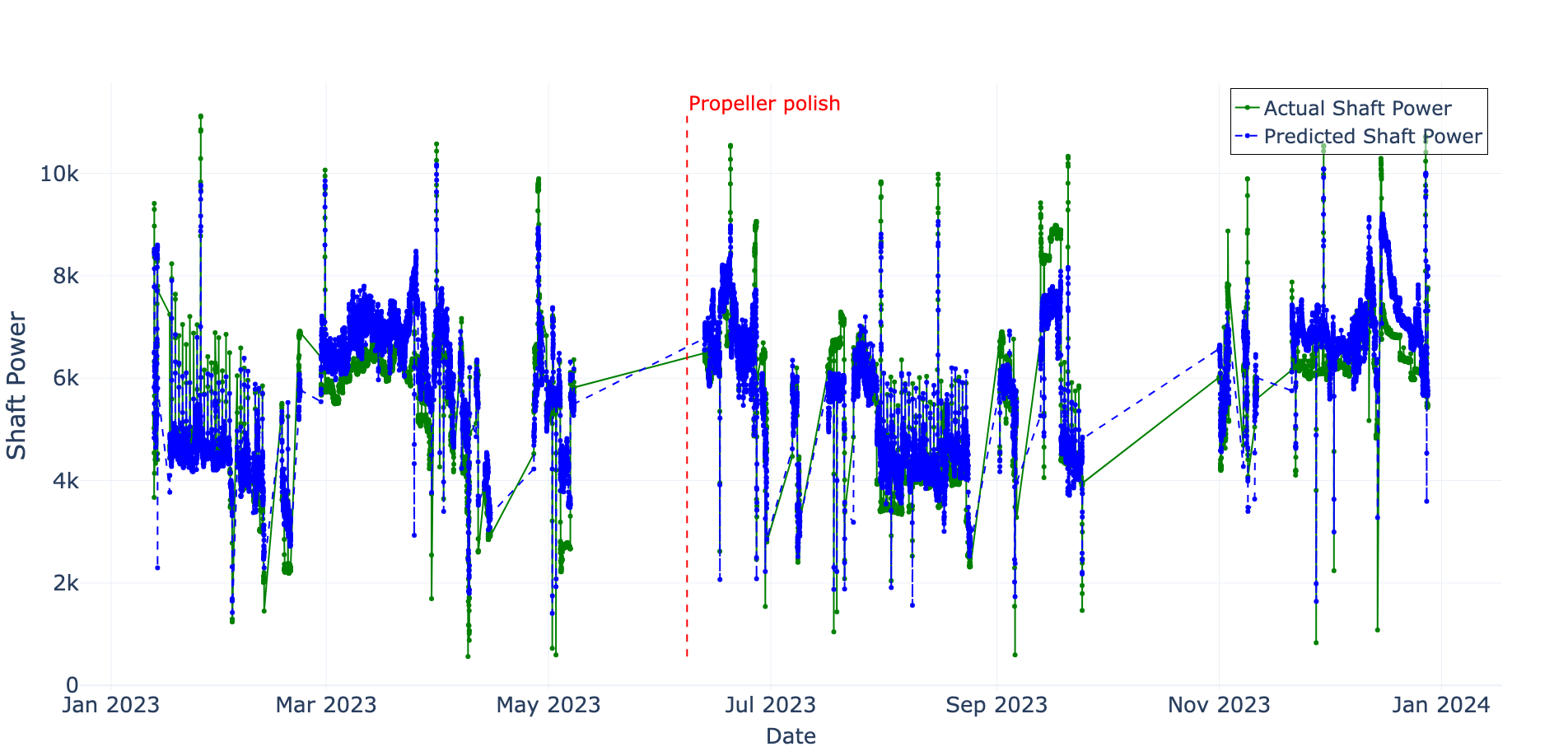}%
\label{fig_vessel_a}}
\hfill
\subfloat[Vessel B]{\includegraphics[width=0.50\textwidth]{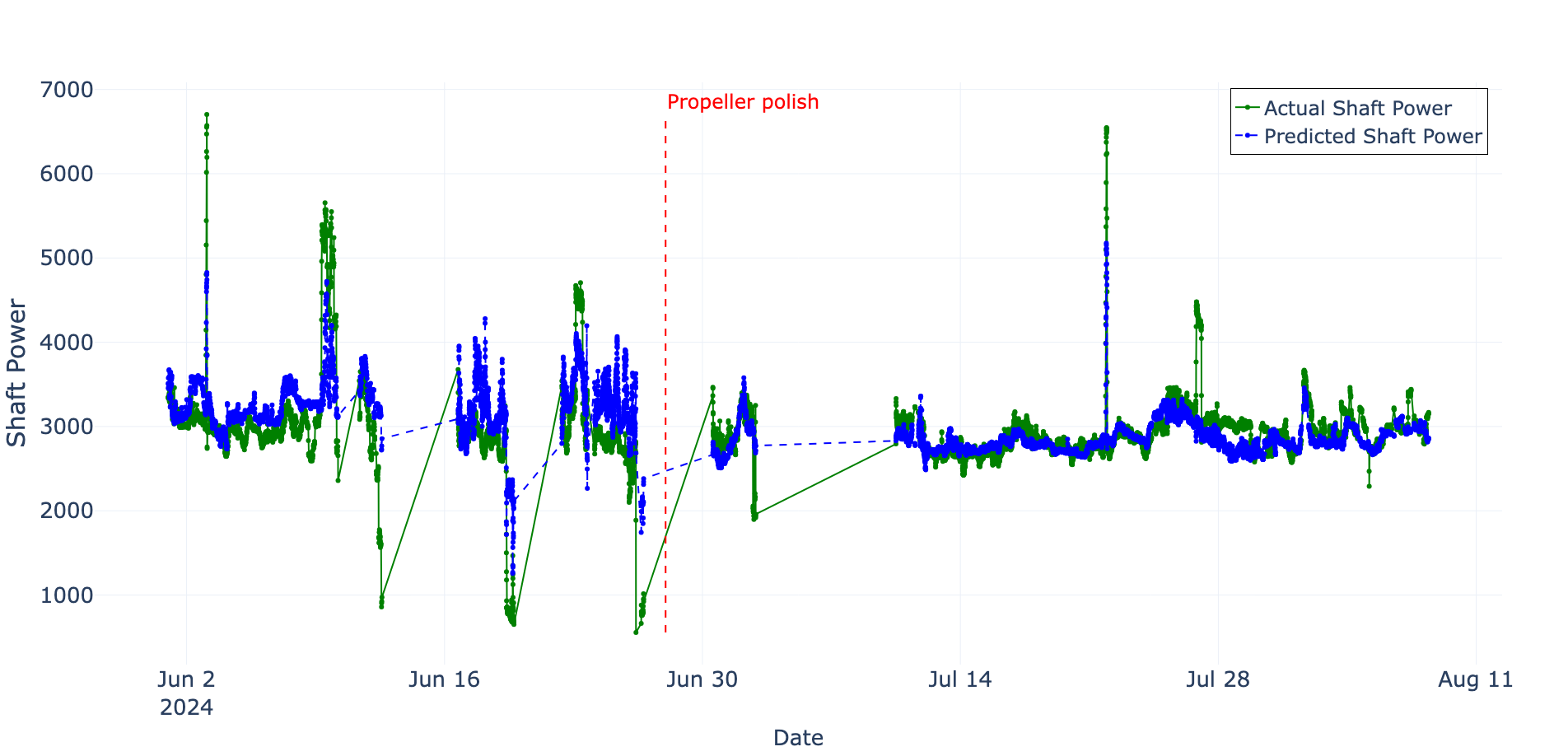}%
\label{fig_vessel_b}}
\vfil
\subfloat[Vessel C]{\includegraphics[width=0.50\textwidth]{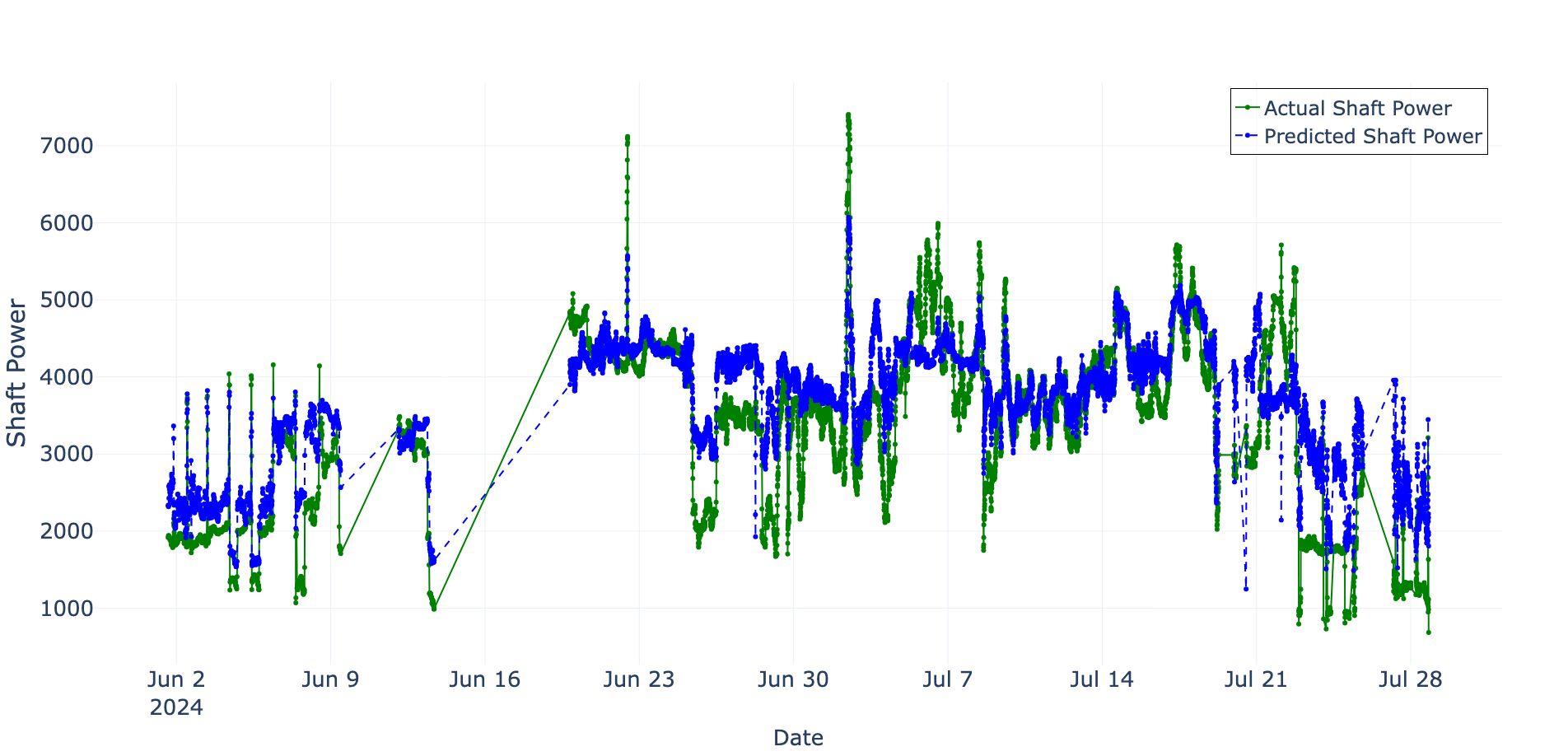}%
\label{fig_vessel_c}}
\hfill
\subfloat[Vessel D]{\includegraphics[width=0.5\textwidth]{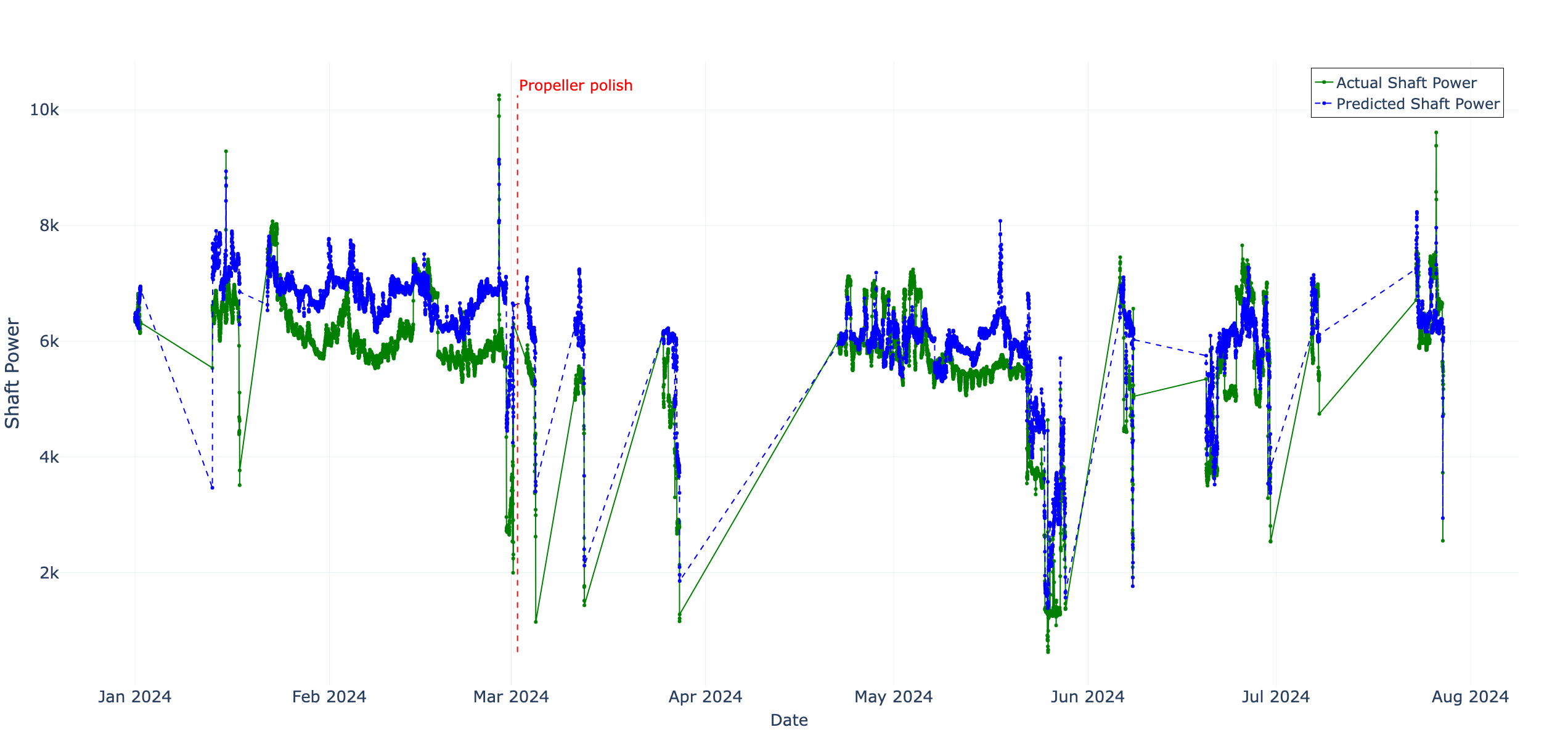}%
\label{fig_vessel_d}}
\caption{Predicted vs. Actual Shaft Power for All Vessels in the Test Set for PGNN.}
\label{fig_performance_over_time}
\end{figure*}

\subsubsection{RQ2 Comparison with Ground Truth}

In this experiment, we present the predictions of the shaft power model on the test sets for each vessel and illustrate how the predictions and ground truth shaft power values (in kW) compare over time. Figure \ref{fig_performance_over_time} shows four plots, each of which corresponds to a vessel that is investigated in this study. In each plot, the x-axis corresponds to the date of the data instance, and the y-axis corresponds to the shaft power. Green represents the actual shaft power, and blue represents the predicted shaft power. Figures \ref{fig_vessel_a}-\ref{fig_vessel_d} depict the model predictions for Vessels A-D, respectively. When the predictions for Vessel A are considered, the model predictions are mostly consistent with the actual shaft power values. Offset patterns are observed between the predicted and actual shaft power from time to time after July 2023 for that vessel. 
For Vessel B, the predictions overlap mainly with the actual shaft power values, especially when the actual shaft power does not change significantly over time (Figure \ref{fig_vessel_b}). For Vessel C, there are also some shifts in the predictions, as expected, for instance, around June 25. Finally, the predictions of Vessel D deviate more from the actual shaft power values, as this vessel is shown to provide higher error scores (MAE and RMSE). On the other hand, after the propeller polish event, the model predictions are more accurate.

\subsubsection{RQ3 Performance under Challenging Weather Conditions}
In this experiment, we evaluate the performance of the proposed method on two ships sailing in the Atlantic Ocean under challenging sea conditions. We use Vessel A as the first ship and another ship of the same size and category as Vessels B-D, which is denoted as Vessel E. The voyage durations are 10 and 13 days for Vessel A and Vessel E, respectively. Particularly, the maximum wave height is 4 and 6 meters for Vessel A and Vessel E. Each voyage is evaluated using models trained on two years of data specific to that ship. Note that in this experiment, the train and test sets for both vessels do not include any dry docking events, and the test period precedes the training set chronologically, as the amount of data collected before those voyages is insufficient for training. Table \ref{tab:dataset-rq3} presents the properties of the dataset used for evaluation in this particular experiment.

\begin{table*}[h]
    \centering
    \caption{Number of instances and data periods for each vessel in training and test sets for the performance under challenging weather conditions.}
    \renewcommand{\arraystretch}{1.5}
    \begin{tabular}{|P{1.5cm}|P{2.4cm}|P{2.2cm}|P{3.4cm}|P{3.4cm}|}
        \rowcolor{gray!20}
        \hline
        \textbf{Vessel} & \textbf{Train Set Instances} & \textbf{Test Set Instances} & \textbf{Train Dates (Start/End)} & \textbf{Test Dates (Start/End)}  \\
        \hline
        Vessel A & 24518 & 1014 & 09-08-2020 / 08-08-2022 & 28-07-2020 / 08-08-2020 \\
        \hline
        Vessel E & 30877 & 971 & 14-12-2020 / 13-12-2022 & 29-11-2020 / 13-12-2020 \\
        \hline
    \end{tabular}
    \label{tab:dataset-rq3}
\end{table*}

\begin{table*}[ht]
    \centering
    \caption{Performance analysis (mean ± std) of different methods under unfavorable sea conditions. \textbf{EF:} Empirical Formulas, \textbf{NN:} Neural Network, \textbf{PGNN:} Physics-guided Neural Network.}
    \renewcommand{\arraystretch}{2.0} 
    \rowcolors{2}{gray!10}{white}
    \begin{tabularx}{\textwidth}{ p{2.0cm} p{1.3cm} >{\centering\arraybackslash}X >{\centering\arraybackslash}X >{\centering\arraybackslash}X >{\centering\arraybackslash}X }
        \toprule
        \rowcolor{gray!30}
        \textbf{Vessel} & \textbf{Method} & \textbf{MAE} & \textbf{RMSE} & \textbf{R\textsuperscript{2}} & \textbf{MAPE (\%)} \\
        \midrule
        
        \multirow{3}{*}{A}
                           & \textbf{EF}    & 1257.02 ± 1.27 & 1308.11 ± 1.99  & -1.48 ± 0.01 & 16.20 ± 0.00 \\
        \textbf{Vessel A}  & \textbf{NN}    & 759.00 ± 133.74 &  928.30 ± 155.79 & -0.27 ± 0.41 & 9.70 ± 1.68 \\
                           & \textbf{PGNN}  & 551.70 ± 63.01 &  664.61 ± 60.65 & 0.35 ± 0.11 & 7.00 ± 0.89  \\
        
        \midrule
        
        \multirow{3}{*}{B}
                          & \textbf{EF}    & 860.33 ± 2.39 & 1075.57 ± 4.32 & 0.36 ± 0.01 & 20.83 ± 0.00 \\
        \textbf{Vessel E} & \textbf{NN}    & 858.58 ± 87.04 & 1066.39 ± 91.28 & 0.37 ± 0.11 & 18.50 ± 1.63 \\
                          & \textbf{PGNN}  & 736.60 ± 32.96 & 969.15 ± 48.17 & 0.48 ± 0.05 & 17.10 ± 0.83 \\   
        \bottomrule
    \end{tabularx}
    \label{tab:atlantic-experiments}
\end{table*}

Table \ref{tab:atlantic-experiments} presents the scores of the models for the metrics, MAE, RMSE, R$^2$ and MAPE, when the vessels operate on the Atlantic Ocean facing challenging sea conditions. For both vessels, PGNN achieves superior results across all reported metrics during the Atlantic crossing voyage. Compared to the results for Vessel A reported in RQ1, improvements are also observed in the MAE, RMSE, and MAPE metrics. It is worth noting that both the training and test sets for this experiment exclude dry-docking events, which may explain the performance gains despite the presence of unfavorable sea conditions. Although Vessel E provides higher error scores than Vessel A for PGNN, when the other baselines are considered, PGNN provides the best scores for that vessel (Vessel E). When the speeds of the vessels are investigated in the case of high waves, it is observed that Vessel E tends to decrease its speed, while Vessel A maintains a constant and higher speed. For both vessels, PGNN provides almost 2\% improvement in MAPE compared to NN on average. 

\section{Discussion and Conclusions}

In this paper, we present a physics-guided neural network-based method for predicting vessel shaft power, where empirical formulas for calm water, waves, and wind resistance are embedded into the neural network design. The presented method is tested on four similar-sized cargo vessels, and compared to the base neural networks without empirical formula guidance and the pure empirical formula-based shaft power method. The results indicate that PGNN achieves promising results, achieving lower error scores than the empirical formula-based shaft power method and the base neural network.

\subsection{Discussion and Limitations}

Traditional approaches, which utilize empirical formulas to predict shaft power, are expected to perform well when data drift cases (i.e., instances of drift due to fouling) are minimal in both the training and test sets. Therefore, when data drift is not present, i.e., fouling is minimal, traditional methods may still yield sufficiently accurate results.

The conducted experiments show that neural network-based approaches outperform the method based on empirical formulas when the vessel undergoes a dry docking event in the training set. Considering the neural network-based approaches, for some vessels, they provide comparable scores, which may be due to both methods using raw weather-related and vessel event-related features as inputs, in addition to the physics formulas. In the experiments, it is observed that PGNN is inclined to significant prediction errors when there is an instant increase or decrease in the shaft power.

The extrapolation capabilities of the neural network-based method are also limited to conditions that the model has not encountered during training \cite{webb2020learning}. For the weather-related conditions, incorporating resistance-based power loss is expected to help PGNN generalize better for unseen test cases during training compared to the base neural network-based method. On the other hand, the physics-guided component of the presented method lacks a representation of fouling; instead, the method incorporates fouling-related features as inputs to the network, rather than including them as a physical loss component. 

\subsection{Future Work}
In future work, we plan to integrate fouling conditions as a part of the physical loss to enhance the learning of the fouling effects on the shaft power of the vessel. Furthermore, an ablation analysis of resistance components used in physical loss is on our agenda. Testing the presented approach on data obtained from different types of vessels is also a potential future direction.

\section*{Data and Code Availability}
The data and code used in the paper are confidential, as they belong to the industrial partner.

\section*{Acknowledgment}
This study has been funded by the Research Council of Norway under grant agreement No. 346603, the GASS project. The study also benefited from the Experimental Infrastructure for Exploration of Exascale Computing (eX3), which is financially supported by the Research Council of Norway under contract number 270053. 

The study has been conducted using E.U. Copernicus Marine Service Information, and the authors acknowledge the use of Copernicus Marine Service data \footnote{https://doi.org/10.48670/moi-00016} \footnote{https://doi.org/10.48670/moi-00017} \footnote{https://doi.org/10.48670/moi-00021}  \footnote{https://doi.org/10.48670/moi-00022}.

The authors thank Joachim Haga for improvements to the empirical resistance formulas and Akriti Sharma and Tetyana Kholodna for helpful discussions on the methodology.

\bibliographystyle{IEEEtran} 
\bibliography{references} 

\end{document}